\documentclass{article}

\usepackage{arxiv}

\usepackage[utf8]{inputenc} 
\usepackage[T1]{fontenc}    
\usepackage{hyperref}       
\usepackage{url}            
\usepackage{booktabs}       
\usepackage{amsfonts}       
\usepackage{nicefrac}       
\usepackage{microtype}      
\usepackage{lipsum}		
\usepackage{graphicx}
\usepackage{natbib}
\usepackage{doi}
\usepackage{listings}

\title{Memori: A Persistent Memory Layer for Efficient, Context-Aware LLM Agents}

\author{
Luiz C. Borro \\
Memori Labs Inc. \\
\texttt{luiz@memorilabs.ai}
\And
Luiz A. B. Macarini \\
Memori Labs Inc. \\
\texttt{lmacarini@memorilabs.ai}
\And
Gordon Tindall \\
Memori Labs Inc. \\
\texttt{gordon@memorilabs.ai}
\And \And
Michael Montero \\
Memori Labs Inc. \\
\texttt{mike@memorilabs.ai}
\And
Adam B. Struck \\
Memori Labs Inc. \\
\texttt{adam@memorilabs.ai}
}

\renewcommand{\shorttitle}{Memori: A Persistent Memory Layer for Efficient, Context-Aware LLM Agents}

\hypersetup{
pdftitle={Memori: A Persistent Memory System for Large Language Models},
pdfsubject={cs.AI, cs.CL, cs.LG},
pdfauthor={Luiz C. Borro},
pdfkeywords={Large Language Models, Memory-Augmented Systems, Context Efficiency, Retrieval-Augmented Generation, Knowledge Representation},
}

\begin{document}
\maketitle

\begin{abstract}
As large language models (LLMs) evolve into autonomous agents, persistent memory at the API layer is essential for enabling context-aware behavior across LLMs and multi-session interactions. Existing approaches force vendor lock-in and rely on injecting large volumes of raw conversation into prompts, leading to high token costs and degraded performance.
We introduce Memori, a LLM-agnostic persistent memory layer that treats memory as a data structuring problem. Its Advanced Augmentation pipeline converts unstructured dialogue into compact semantic triples and conversation summaries, enabling precise retrieval and coherent reasoning.
Evaluated on the LoCoMo benchmark, Memori achieves 81.95\% accuracy, outperforming existing memory systems while using only 1,294 tokens per query ($\sim$5\% of full context). This results in substantial cost reductions, including 67\% fewer tokens than competing approaches and over 20$\times$ savings compared to full-context methods.
These results show that effective memory in LLM agents depends on structured representations instead of larger context windows, enabling scalable and cost-efficient deployment.

Code: \url{https://github.com/MemoriLabs/Memori}
\end{abstract}


\section{Introduction}

Large language models (LLMs) have quickly become sophisticated AI agents. These foundation-model-powered systems perform well in research, software engineering, and scientific discovery, driving the move toward general intelligence  \citep{hu2025memory}. Modern agents now go beyond using only LLMs by adding reasoning, planning, perception, memory, and tool use \citep{xi2025rise}. These components let LLMs act as adaptive systems that interact with their environments and improve over time.

Among these capabilities, memory stands out as a foundational pillar. Unlike reasoning or tool use, which are increasingly internalized within model parameters, memory remains largely dependent on external system design. This dependency arises because LLM parameters cannot be updated in real time during deployment \citep{shinn2023reflexion}. Memory mechanisms, therefore, play a key role in enabling agents to persist information across interactions, adapt to user context, and evolve based on experience \citep{packer2023memgpt}.

This reliance on external memory is especially apparent from an application perspective: persistent memory is essential \citep{hu2025memory}. Domains such as personalized assistants, recommendation systems, social simulations, and complex investigative workflows all require agents to retain and reason over historical information \citep{zhong2024memorybank, hu2025memory}. Without memory, these systems behave as stateless responders, repeatedly reprocessing context and failing to build continuity over time \citep{packer2023memgpt, wang2024survey}. From a broader research perspective, agents’ ability to continually evolve through interaction is central to the pursuit of general intelligence. This capacity is fundamentally grounded in memory.

Enabling long-term, cross-session, cross-model memory introduces significant challenges. Naively storing and injecting past interactions into the prompt leads to rapidly growing context windows. This increases both cost and instability. As context size grows, models become more prone to overlooking critical information. They may produce inconsistent outputs and suffer from what is commonly referred to as \textit{context rot}, in which relevant information is present but not effectively used \citep{hong2025context}.

These limitations highlight a key insight: memory in LLM systems is not simply a storage problem, but a structuring problem. The challenge is to transform noisy, unstructured conversational data into representations that are efficient to retrieve. These representations must also be effective for downstream reasoning.

Memori implements this as a persistent memory layer that incrementally distills conversational data into structured representations. This process is handled by Advanced Augmentation, a memory creation pipeline that extracts, compresses, and organizes high-signal information from raw interactions for efficient retrieval and downstream use. Through empirical evaluation on the LoCoMo benchmark, we demonstrate that high-quality memory structuring enables strong reasoning performance while  reducing the number of tokens required in the prompt, thereby improving the cost-efficiency and scalability of LLM agents.

\section{System Architecture}
\begin{figure}
	\centering
    \includegraphics[width=0.8\linewidth]
    {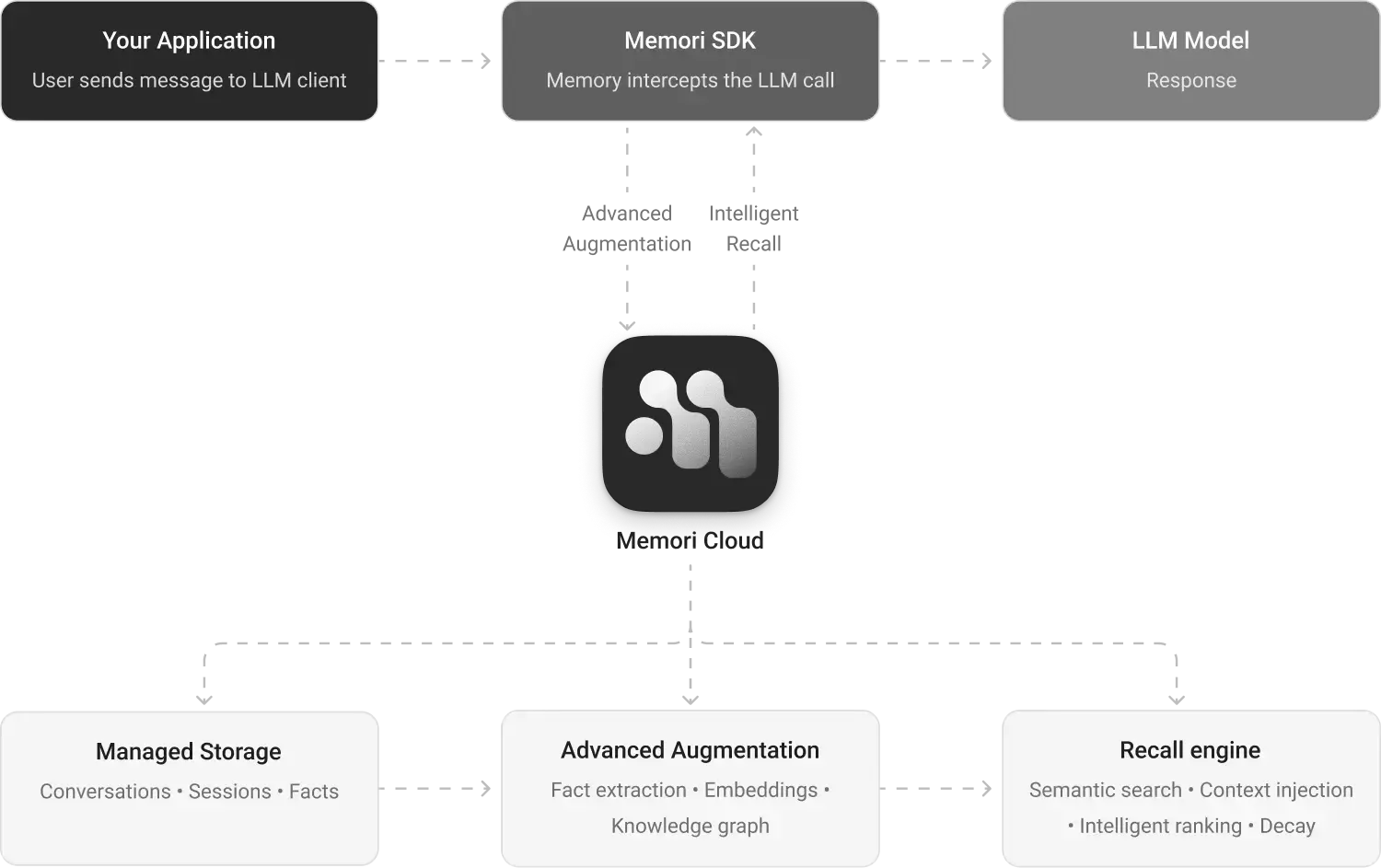}
	\caption{High-level architecture illustrating the system’s structure, data flow, and the interaction between its core components.}
	\label{fig:fig1}
\end{figure}

As depicted in Figure \ref{fig:fig1}, Memori operates as a decoupled memory layer positioned between the application logic and the underlying LLM. The system integrates via a lightweight Memori SDK, seamlessly wrapping existing LLM clients to intercept requests and manage memory natively.

The core differentiators of the Memori architecture lie in how it structures unstructured data and how it intelligently retrieves that data for reasoning.

\subsection{Advanced Augmentation: Structuring the Unstructured}
Raw conversation logs are noisy, filled with colloquialisms, pleasantries, self-corrections, and tangential discussions. When these raw, unstructured transcripts are directly chunked and embedded, as is standard in traditional RAG architectures, the resulting vector space becomes heavily cluttered. Direct retrieval from this noisy data is highly inefficient, leading to false positives, contradictory context, and massively inflated token consumption during the generation phase.

To solve this, Advanced Augmentation functions as an automated cognitive filter. It is a background memory creation pipeline designed to distill raw dialogue into searchable memory assets, shifting the system's memory from mere text storage to an organized knowledge base.

\begin{itemize}
\item \textbf{Semantic Extraction \& Triple Generation:} Rather than saving sentences, the pipeline deconstructs dialogue messages into atomic units of knowledge. It actively scans conversations for concrete facts, user preferences, constraints, and evolving attributes, structuring them into semantic triples (subject–predicate–object). Each triple is then linked to the exact conversation in which it was mentioned. This design delivers two key advantages. First, it produces a low-noise, high-signal index that improves vector search retrieval accuracy. Second, it functions as a compression layer.

\item \textbf{Conversation Summarization:} While semantic triples excel at capturing granular, static facts, they inherently strip away the surrounding context. An isolated triple might state what a user prefers, but it lacks the narrative of why a decision was made or how a user’s goal evolved throughout a specific interaction. To bridge this gap, the pipeline simultaneously generates Conversation Summaries. These are concise, high-level overviews of specific conversational threads that capture the user’s overarching intent, the dialogue's chronological progression, and the task's implicit context. Because triples are tied to their source, each individual triple can be directly linked to the proper summary of the conversation in which it appears, allowing the system to easily retrieve the background story behind any isolated fact.
\end{itemize}

Advanced Augmentation creates an interconnected, dual-layered memory asset: Triples provide the precise, token-efficient facts needed for exact recall, while Conversation Summaries provide the cohesive narrative flow required for the LLM to understand temporal changes and execute complex reasoning. By linking atomic triples directly to the summaries of the conversations they originated from, the system ensures that granular facts are never divorced from their broader context.

\section{Experiments}
The primary objective of these experiments is to evaluate the quality and accuracy of the memory assets produced by Memori's Advanced Augmentation pipeline.

\subsection{Dataset: The LoCoMo Benchmark}
The primary dataset utilized for benchmarking is the Long Conversation Memory (LoCoMo) dataset \citep{maharana2024locomo}. LoCoMo is a rigorous framework engineered to evaluate an AI agent's ability to track, retain, and synthesize information across extensive, multi-session chat histories. Unlike standard QA datasets, LoCoMo challenges models with complex state tracking, temporal reasoning, and the retrieval of subtle user preferences buried deep within noisy, unstructured conversational logs. 

The category alignment and question distribution are detailed in Table \ref{tab:locomo_categories}, presented in Appendix ~\ref{appendix_c}.

To ensure a fair comparison with other published results on this benchmark, we excluded the adversarial category from the evaluation \citep{chhikara2025mem0, du2025memr}.

\subsection{Evaluating Memory Extraction via Advanced Augmentation}
To measure the quality of Memori's Advanced Augmentation, all sections of each LoCoMo conversation were processed through the pipeline. Each session produced a set of semantic triples along with conversation-level summaries. The extracted triples were embedded using the Gemma-300 embedding model, enabling efficient semantic retrieval for the benchmark’s question-answering tasks. All generated memories were indexed and stored locally using FAISS to support fast similarity search. The ultimate accuracy of the LLM’s answers serves as a direct reflection of how well the Advanced Augmentation pipeline structured, preserved, and surfaced the relevant facts.

\subsection{Answer Generation}
Each question in the LoCoMo benchmark was answered using GPT-4.1-mini, conditioned on the retrieved triples and their corresponding summaries (the utilized prompt is presented in the Appendix \ref{appendix_a}). Triples were retrieved using a hybrid search approach that combines cosine similarity over embeddings with BM25 keyword matching.

\subsection{Performance Metrics: LLM-as-a-Judge}
We employ an LLM-as-a-Judge methodology (the utilized prompt is presented in the Appendix \ref{appendix_b}), using GPT-4.1-mini as the evaluator. The judge model analyzes the user query, the ground-truth answer, and the generated response to provide a nuanced assessment. 

\subsection{Token-Driven Cost Analysis}
Beyond response quality, practical deployment considerations are paramount for enterprise AI applications. We evaluate Memori against traditional architectures (e.g., standard RAG) by systematically measuring system efficiency as a function of context consumption. 

The absolute number of tokens added to the LLM prompt is the primary driver of operational costs in conversational AI. We measure the exact number of tokens extracted during retrieval and injected into the prompt context. This metric highlights a critical architectural distinction: while traditional architectures consume massive token budgets by indiscriminately injecting large, raw text chunks or full histories into the prompt, Memori retrieves highly concise, structured memory facts. By minimizing the context footprint, Memori directly curtails API expenditure and optimizes operational economics.

\subsection{Results and Analysis}
This section summarizes how Memori’s Advanced Augmentation performed on the LoCoMo benchmark. We compare Memori against established memory systems, including Zep\footnote{\url{https://github.com/getzep/zep}}, LangMem\footnote{\url{https://github.com/langchain-ai/langmem}}, and Mem0\footnote{\url{https://github.com/mem0ai/mem0}}, using their official open-source implementations. Using our LLM-as-a-Judge framework, we evaluated four reasoning categories (Multi-Hop, Temporal, Open-Domain, and Single-Hop) and compared Memori against several memory baselines and a Full-Context ceiling. We also examined the vital tradeoff between output accuracy and token cost efficiency. The results are presented in Table \ref{tab:llm_judge_locomo}.

\begin{table}[h]
    \centering
    \caption{LLM-as-a-Judge Evaluation Results on the LoCoMo Benchmark. 
    This table compares the factual accuracy and reasoning capabilities of Memori’s Advanced Augmentation assets against state-of-the-art baselines and a full-context ceiling. Memori performance values were computed using the average of three rounds. Results for Mem0, Zep, LangMem and Full-context were retrieved from \cite{du2025memr}.}
    \begin{tabular}{lccccc}
        \toprule
        \textbf{Method} & \textbf{Single-hop (\%)} & \textbf{Multi-hop (\%)} & \textbf{Open-domain (\%)} & \textbf{Temporal (\%)} & \textbf{Overall (\%)} \\
        \midrule
        Memori & 87.87 & 72.70 & 63.54 & 80.37 & 81.95 \\
        Zep & 79.43 & 69.16 & 73.96 & 83.33 & 79.09 \\
        LangMem & 74.47 & 61.06 & 67.71 & 86.92 & 78.05 \\
        Mem0 & 62.41 & 57.32 & 44.79 & 66.47 & 62.47 \\
        Full-Context (Ceiling) & 88.53 & 77.70 & 71.88 & 92.70 & 87.52 \\
        \bottomrule
    \end{tabular}
    \label{tab:llm_judge_locomo}
\end{table}

\begin{figure}
	\centering
    \includegraphics[width=0.8\linewidth]
    {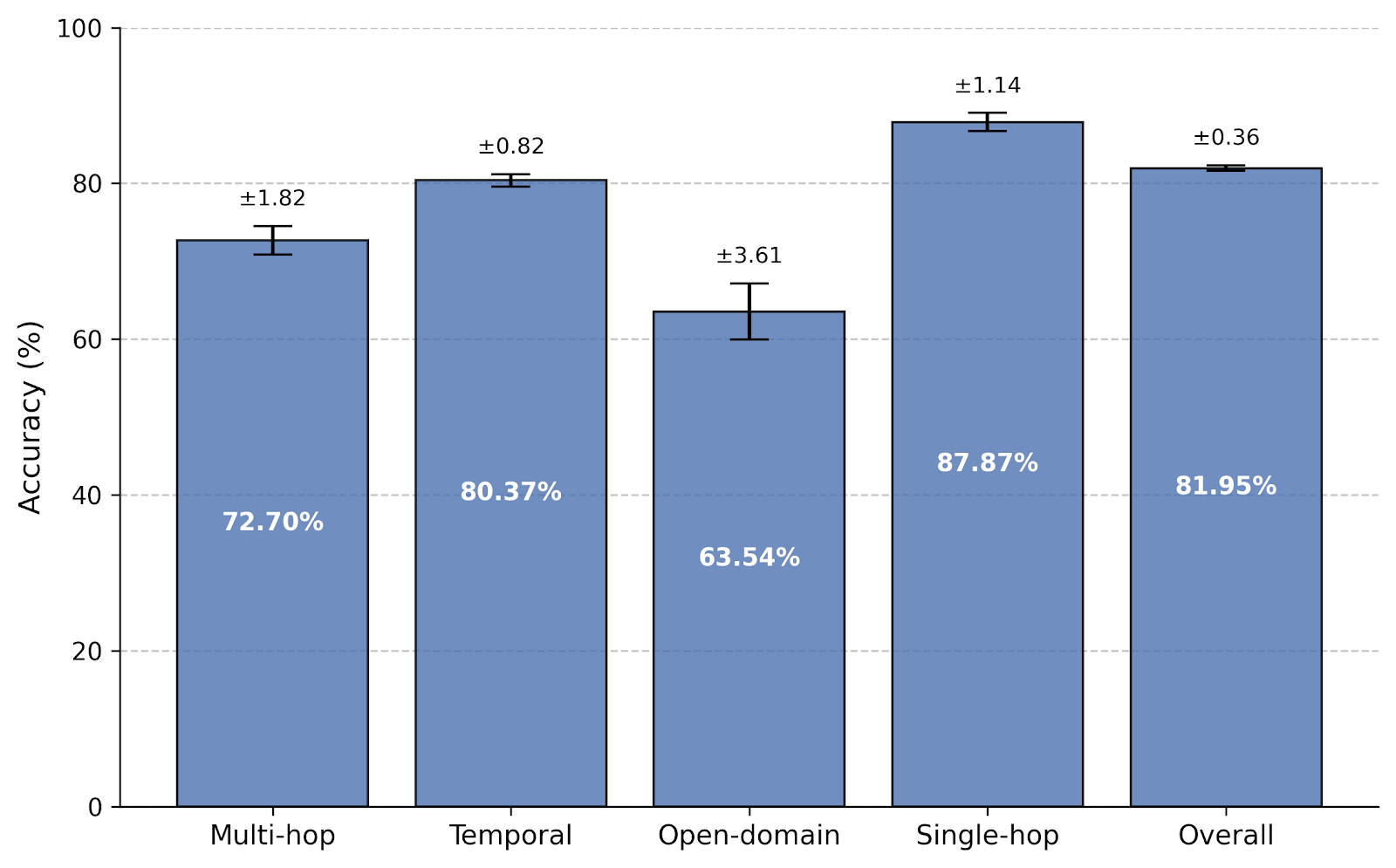}
	\caption{Accuracy of Memori across different reasoning categories. Bar heights represent the mean accuracy (\%), while error bars indicate the standard deviation for $n=3$ runs.}
	\label{fig:fig2}
\end{figure}

Graphical representation of the Memori's average accuracy along with the standard deviation is presented in Figure \ref{fig:fig2}.

\subsection{Overall Performance}
As expected, the Full-Context setup achieved the highest score (87.52\%). However, passing the entire conversation history into the prompt is fundamentally impractical in production due to prohibitive token costs, uncontrolled context expansion, and context degradation over time.

Among retrieval-based systems, Memori achieved a leading overall score of 81.95\%, successfully outperforming Zep (79.09\%), LangMem (78.05\%), and Mem0 (62.47\%). This validates the assumption that by structuring unstructured chat logs into semantic triples and their summaries, Memori effectively isolates high-signal knowledge. This structured memory design significantly narrows the gap to the Full-Context ceiling while keeping context windows highly manageable and operational token costs low.

The reported overall scores for all methods were computed as a weighted average based on the number of questions in each LoCoMo category, as detailed in Table \ref{tab:locomo_categories}, at Appendix \ref{appendix_c}.

\subsection{Performance by Category}
Analyzing the results across reasoning categories highlights the specific strengths of Memori's extracted memory assets:

\begin{itemize}
\item \textbf{Single-Hop Reasoning (87.87\%):} Memori excels in direct fact retrieval, outperforming both LangMem (74.47\%) and Zep (79.43\%). By minimizing conversational noise and structuring data cleanly, the LLM is fed exact, undeniable facts, minimizing token consumption while maximizing direct recall.

\item \textbf{Temporal Reasoning (80.37\%):} Memori outperforms Mem0 (66.47\%) but trails LangMem (86.92\%) and Zep (83.33\%) in temporal tracking. Isolated semantic triples capture static facts but often miss the temporal context needed to identify changes in user states or preferences across sessions. Memori’s summaries help rebuild this timeline, but the results show it needs better temporal reasoning.

\item \textbf{Multi-Hop Reasoning (72.70\%):} Memori performs strongly when asked to connect disparate pieces of information, outperforming Zep (69.16\%) and trailing LangMem (61.06\%) by a narrow margin. The combination of precise triples and cohesive summaries provides the necessary backdrop that helps the LLM connect isolated facts without needing the entire conversational transcript injected into the prompt.

\item \textbf{Open-Domain Reasoning (63.54\%):} This category remains challenging across all retrieval-based systems. Open-ended questions often lack clear retrieval anchors, making them difficult to match with granular triples. Furthermore, these queries require broad synthesis across massive contexts rather than simple fact extraction. While Memori lags slightly behind LangMem (67.71\%) here, it is important to note that improving open-domain scores typically requires retrieving significantly larger chunks of text, which actively works against the system's core operational goal: strictly minimizing the number of tokens added to the context to control API costs.
\end{itemize}

\subsection{Token Usage and Cost Efficiency}
Traditional memory architectures and standard RAG setups often rely on retrieving raw, uncompressed text chunks. This indiscriminately injects conversational noise and redundant dialogue into the prompt, consuming context limits and inflating API bills.

Memori’s Advanced Augmentation pipeline completely bypasses this inefficiency by acting as an intelligent cognitive filter. Rather than retrieving raw text, it compresses chat logs into structured representations, including dense semantic triples and concise conversation-level summaries.  As a result, only high-signal, structured information is passed to the LLM, minimizing context overhead while preserving relevant context. 

\begin{table}[h]
    \centering
    \caption{Token Usage and Cost Efficiency. 
    This table analyzes the operational efficiency of each method by measuring the absolute number of tokens added to the context and the resulting cost per query. Costs are computed based on current \textit{gpt-4.1-mini} pricing: \$0.8 per 1M tokens. Mem0, Zep and Full-Context values were retrieved from \cite{chhikara2025mem0}.}
    \begin{tabular}{lccc}
        \toprule
        \textbf{Method} & \textbf{Added Tokens to Context (mean)} & \textbf{Context Cost (\$)} & \textbf{Context Footprint (\%)} \\
        \midrule
        Memori & 1,294 & 0.001035 & 4.97 \\
        Full-Context & 26,031 & 0.020825 & 100.00 \\
        Mem0 & 1,764 & 0.001411 & 6.78 \\
        Zep & 3,911 & 0.003129 & 15.02 \\
        \bottomrule
    \end{tabular}
    \label{tab:token_efficiency}
\end{table}

Memori requires an average of only 1,294 tokens to ground each LLM response. This token footprint represents just 4.97\% of the full conversational context, while achieving the 81.95\% overall accuracy detailed in the previous section.
When compared to competing memory frameworks, the operational advantages become even more pronounced:

\begin{itemize}
    \item C\textbf{ompared to Zep (3,911 tokens):} Memori reduces the prompt size by roughly 67\% per query, directly cutting API inference costs by the same margin, while simultaneously delivering a higher accuracy score (81.95\% vs. 79.09\%).

    \item \textbf{Compared to the Full-Context approach (26,031 tokens):} Passing the entire history is financially unsustainable for persistent agents, costing over 20 times more per turn than Memori. Furthermore, repeatedly injecting 26K+ tokens drastically increases the risk of "lost in the middle" hallucinations.
\end{itemize}

\section{Conclusion}
For LLM agents to scale in production, persistent memory must address two fundamental challenges: context degradation and rapidly increasing token costs. Memori approaches this not as a storage issue, but as a data structuring problem.

Through its Advanced Augmentation memory creation pipeline, Memori transforms noisy conversational logs into compact, high-signal representations, combining precise semantic triples with coherent conversation-level summaries. This dual representation enables accurate fact retrieval alongside strong temporal and contextual reasoning, without inflating the prompt with unnecessary tokens.

Our evaluation on the LoCoMo benchmark demonstrates the effectiveness of this approach:

\begin{itemize}
    \item \textbf{High-Quality Reasoning:} Memori achieves state-of-the-art performance among retrieval-based systems, with particularly strong gains in temporal and single-hop reasoning - highlighting the impact of structured memory on reasoning fidelity.

    \item \textbf{Minimal Context Footprint:} Responses are grounded using a small fraction of the original conversation, showing that well-structured memory can replace large, unfiltered context without sacrificing accuracy.

    \item \textbf{Cost-Efficient Scaling:} By significantly reducing the number of tokens injected into the prompt, Memori directly lowers inference costs and enables sustainable deployment of long-running agents.
\end{itemize}

These results highlight a fundamental shift in memory design for LLM systems: performance is determined not by how much context is used, but by the quality of its structure. 

Memori eliminates the traditional tradeoff between reasoning quality and operational cost. By delivering accurate, cross-session recall with a compact context footprint, it provides a practical and scalable foundation for deploying persistent AI agents in real-world environments.

\newpage
\appendix

\section{Appendix A}\label{appendix_a}
Prompt Template for Results Generation \citep{chhikara2025mem0}.
\begin{lstlisting}[basicstyle=\ttfamily\small, breaklines=true]
ANSWER_PROMPT = """
    You are an intelligent memory assistant tasked with retrieving accurate information from conversation memories.

    # CONTEXT:
    You have access to two types of information from a conversation:
    - Memories: timestamped factual triples extracted from conversations.
    - Summaries: high-level conversation summaries (also timestamped) that
      provide broader context around the memories.

    # INSTRUCTIONS:
    1. Carefully analyze all provided memories and summaries
    2. Pay special attention to the timestamps to determine the answer
    3. If the question asks about a specific event or fact, look for direct evidence in the memories
    4. If the memories contain contradictory information, prioritize the most recent memory
    5. If there is a question about time references (like "last year", "two months ago", etc.),
       calculate the actual date based on the memory timestamp. For example, if a memory from
       4 May 2022 mentions "went to India last year," then the trip occurred in 2021.
    6. Always convert relative time references to specific dates, months, or years. For example,
       convert "last year" to "2022" or "two months ago" to "March 2023" based on the memory
       timestamp. Ignore the reference while answering the question.
    7. Focus only on the content of the memories. Do not confuse character
       names mentioned in memories with the actual users who created those memories.
    8. The answer should be less than 5-6 words.

    # APPROACH (Think step by step):
    1. First, examine all memories that contain information related to the question
    2. Use summaries for broader context when memories alone are insufficient
    3. Examine the timestamps and content carefully
    4. Look for explicit mentions of dates, times, locations, or events that answer the question
    5. If the answer requires calculation (e.g., converting relative time references), show your work
    6. Formulate a precise, concise answer based solely on the evidence in the memories
    7. Double-check that your answer directly addresses the question asked
    8. Ensure your final answer is specific and avoids vague time references

    {{memories}}

    Question: {{question}}
    Answer:
    """
\end{lstlisting}

\newpage
\section{Appendix B}\label{appendix_b}
Prompt Template for LLM-as-a-Judge \citep{chhikara2025mem0}.

\begin{lstlisting}[basicstyle=\ttfamily\small, breaklines=true]
ACCURACY_PROMPT = """
Your task is to label an answer to a question as 'CORRECT' or 'WRONG'. You will be given the following data:
    (1) a question (posed by one user to another user),
    (2) a 'gold' (ground truth) answer,
    (3) a generated answer
which you will score as CORRECT/WRONG.

The point of the question is to ask about something one user should know about the other user based on their prior conversations.
The gold answer will usually be a concise and short answer that includes the referenced topic, for example:
Question: Do you remember what I got the last time I went to Hawaii?
Gold answer: A shell necklace
The generated answer might be much longer, but you should be generous with your grading - as long as it touches on the same topic as the gold answer, it should be counted as CORRECT.

For time related questions, the gold answer will be a specific date, month, year, etc. The generated answer might be much longer or use relative time references (like "last Tuesday" or "next month"), but you should be generous with your grading - as long as it refers to the same date or time period as the gold answer, it should be counted as CORRECT. Even if the format differs (e.g., "May 7th" vs "7 May"), consider it CORRECT if it's the same date.

Now it's time for the real question:
Question: {question}
Gold answer: {gold_answer}
Generated answer: {generated_answer}

First, provide a short (one sentence) explanation of your reasoning, then finish with CORRECT or WRONG.
Do NOT include both CORRECT and WRONG in your response, or it will break the evaluation script.

Just return the label CORRECT or WRONG in a json format with the key as "label".
"""
\end{lstlisting}

\newpage
\section{Appendix C}\label{appendix_c}
This table presents the mapping between question categories in the LoCoMo dataset and their corresponding evaluation order, along with the number of questions in each category.

\begin{table}[h]
    \centering
    \caption{The alignment of the orders and categories in the LoCoMo dataset.}
    \begin{tabular}{lccccc}
        \toprule
        \textbf{} & \textbf{Multi-Hop} & \textbf{Temporal} & \textbf{Open-Domain} & \textbf{Single-Hop} & \textbf{Adversarial} \\
        \midrule
        \textbf{Order} & Category 1 & Category 2 & Category 3 & Category 4 & Category 5 \\
        \textbf{\# Questions} & 282 & 321 & 96 & 830 & 445 \\
        \bottomrule
    \end{tabular}
    \label{tab:locomo_categories}
\end{table}

\bibliographystyle{unsrtnat}
\bibliography{references}  






\end{document}